\documentclass{article}

\usepackage[final]{corl_2026} 
\makeatletter
\renewcommand{\@conferencelocation}{Austin, Texas, USA}
\makeatother
\usepackage[table]{xcolor}
\usepackage[utf8]{inputenc}
\usepackage[T1]{fontenc}
\usepackage{amsmath}
\usepackage{amsfonts}
\usepackage{booktabs}
\usepackage{float}
\usepackage{graphicx}
\usepackage{enumitem}
\usepackage{microtype}
\usepackage{xcolor}
\usepackage{amssymb}
\title{ALIGN: Aligning Condition Estimation and Expression in Advantage-Conditioned Behavior Cloning}

\author{
  \textbf{Zechen Wang}\textsuperscript{3*}\quad
  \textbf{Zhenqi Qiu}\textsuperscript{1,3*}\quad
  \textbf{Houlong Xiong}\textsuperscript{2,3*}\quad
  \textbf{Suohang Zhang}\textsuperscript{4}\\
  \textbf{Yiyu Ren}\textsuperscript{3}\quad
  \textbf{Wanting Xu}\textsuperscript{1}\quad
  \textbf{Hongfei Niu}\textsuperscript{3}\quad
  \textbf{Chengyang He}\textsuperscript{5}\\
  \textbf{Ge Sun}\textsuperscript{5\textdagger}\quad
  \textbf{Ran Cheng}\textsuperscript{3}\quad
  \textbf{Qian Zhu}\textsuperscript{3\textdagger}\\[0.5em]
  \normalfont\textsuperscript{1} ShanghaiTech University\quad
  \textsuperscript{2} Shanghai Jiao Tong University\\
  \normalfont\textsuperscript{3} PrimeBot\quad
  \textsuperscript{4} Zhejiang University\\
  \normalfont\textsuperscript{5} National University of Singapore\\
  \normalfont\textsuperscript{*} Equal contribution;\quad
  \textsuperscript{\textdagger} Corresponding authors
}

\hypersetup{
  pdftitle={ALIGN: Aligning Condition Estimation and Expression in Advantage-Conditioned Behavior Cloning},
  pdfauthor={Zechen Wang, Zhenqi Qiu, Houlong Xiong, Suohang Zhang, Yiyu Ren, Wanting Xu, Hongfei Niu, Chengyang He, Ge Sun, Ran Cheng, Qian Zhu}
}

\begin{document}
\raggedbottom
\maketitle
\vspace{-0.3in}

\begin{abstract}
Human-collected robot manipulation data often mix successful behavior with retries, pauses, recoveries, and failures. \textbf{A}dvantage-\textbf{C}onditioned \textbf{B}ehavior \textbf{C}loning (AC-BC) can exploit such mixed-quality data, but only if the conditioning signal is both accurately constructed from offline trajectories and effectively incorporated into the policy's action-vector-field prediction. 
Existing pipelines can weaken this signal by thresholding estimator-derived advantage scores into noisy conditions, and by routing conditions through prompts that weakly affect the action vector field. 
We introduce ALIGN, a condition-aligned AC-BC framework for vision-language-action policies. 
ALIGN directly predicts categorical pairwise conditions in the policy-facing label space and injects them into a DiT action expert through AdaRMS modulation fused with flow time. 
We evaluate ALIGN on deliberately mixed real-robot datasets for ARX X5 bowl stacking and Franka key hanging, each combining standard demonstrations with retry-heavy trajectories. 
ALIGN improves condition fidelity, strengthens conditional vector-field separability, and achieves 90.0\% and 56.7\% closed-loop success, compared with 56.7\% and 33.3\% for the strongest AC-BC baseline.
These results indicate that aligning conditions from estimation to action prediction is an effective way to extract robust manipulation policies from data in which success,  recovery and error coexist.
\end{abstract}
\begin{figure}[H]
  \centering
  \includegraphics[width=0.98\linewidth]{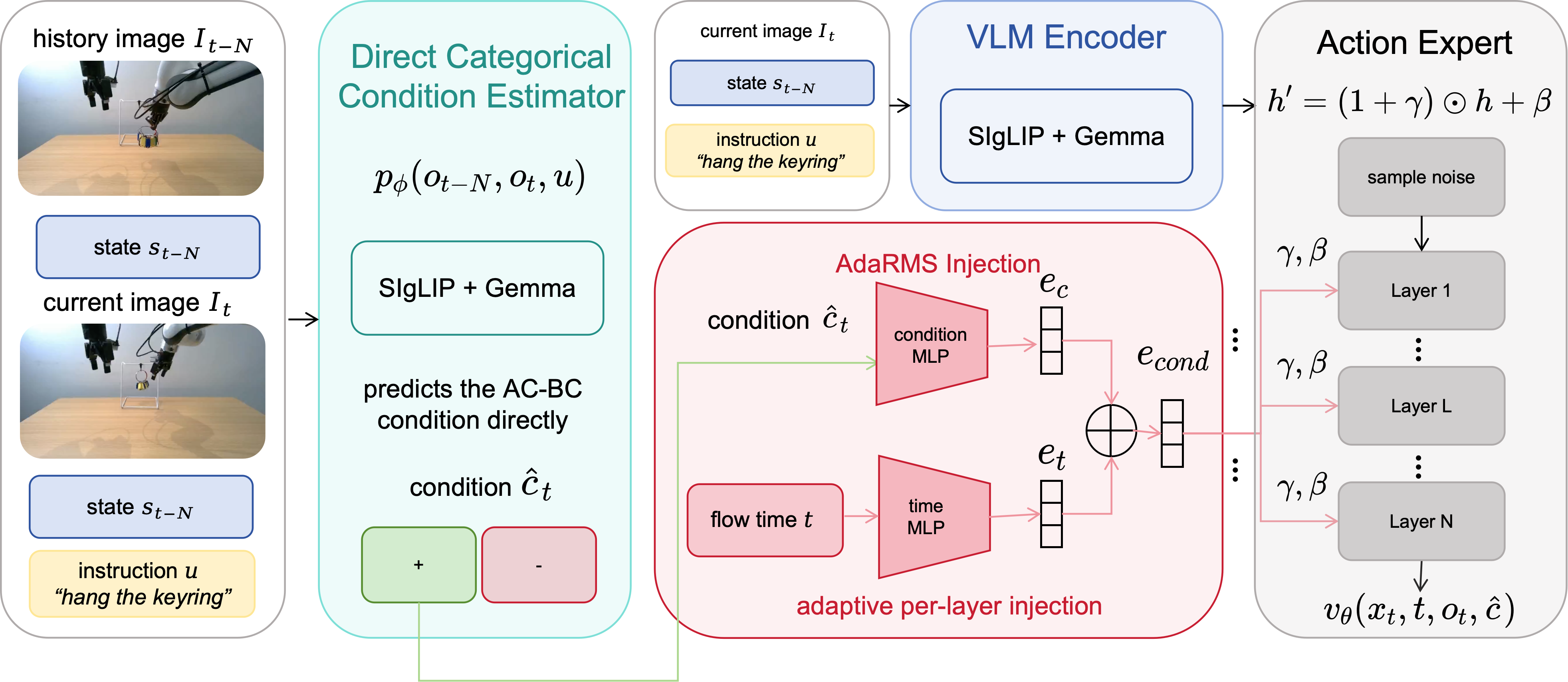}
  \vspace{-0.55em}
  \caption{Overview of ALIGN. Left: condition estimation. Given a historical observation, the current observation, robot states, and the task instruction, the direct categorical estimator predicts the AC-BC condition \(\hat{c}_t\) from an ordered observation pair. Right: AC-BC policy training with estimated conditions. The current observation stream is encoded by the VLM, while AdaRMS combines \(\hat{c}_t\) with the flow-time embedding to generate per-layer modulation parameters for the DiT action expert, so the predicted action vector field is explicitly conditioned on the estimated training condition.}
  \vspace{-1.0em}
  \label{fig:architecture}
\end{figure}


\section{Introduction}

Robot manipulation datasets collected from human teleoperation are rarely clean: besides successful execution, they contain pauses, retries, recoveries, and erroneous actions. This heterogeneity is important when adapting vision-language-action (VLA) models, which have emerged as strong robot policy backbones~\citep{zitkovich2023rt,kim2024openvla,intelligence2025pi_,team2025gemini,team2025gemini15}, including diffusion- and flow-based variants for multimodal action distributions~\citep{chi2025diffusion,black2024pi_0}. 
\textbf{A}dvantage-\textbf{C}onditioned \textbf{B}ehavior \textbf{C}loning (AC-BC) offers a natural way to extract useful behavior from such mixed-quality offline data~\citep{peng2019advantage,ashvin2020accelerating,intelligence2025pi06vlalearnsexperience}: it estimates behavior quality, converts it into a policy-facing condition, trains imitation conditioned on that label, and deploys the policy under the desired high-quality condition. 
In principle, conditioning can retain useful recovery behavior while learning from both positive- and negative-conditioned examples. Filtering with the same classifier can also retain recovery frames predicted as positive; the distinction is that filtering discards predicted-negative examples. 
In practice, its benefit depends on whether the condition is both correctly constructed from data and strongly expressed by the policy.

We argue that existing VLA-based AC-BC pipelines can lose this conditioning signal at both stages. First, the condition is often constructed indirectly: a continuous value, progress, or reward score is evaluated at two frames, differenced, thresholded, and then mapped to a discrete positive/negative label~\citep{intelligence2025pi06vlalearnsexperience,yu2026chi_,chen2025sarm}. This indirect conversion is fragile: continuous value/progress estimates are imperfect predictions, differencing can amplify local prediction noise, and thresholding turns near-boundary errors into discrete condition flips. Our temporal-gap supervision likewise represents local task progress, rather than independently annotated action quality or calibrated advantage. We study how this progress-based signal is converted into the condition consumed by AC-BC. The supervision used to define this signal is a separate design axis from estimator parameterization and condition expression. Second, even a correct condition may be weakly expressed. Prompt-based conditioning is convenient for VLA policies, but the condition must pass through the language or vision-language context before reaching the flow-based action expert. If changing the condition barely changes the predicted action vector field, the learned policy has little ability to act differently under different conditions.

We introduce ALIGN, an AC-BC framework that aligns the policy-facing condition from estimation to action prediction. For condition construction, ALIGN replaces post-hoc thresholding of scalar value/progress differences with a direct categorical pairwise estimator, $p_\phi(c \mid o_i, o_j)$, trained to predict the same discrete condition consumed by the policy. 
For condition expression, ALIGN fuses the categorical condition with the flow-time embedding and injects it into the DiT action expert through AdaRMS modulation, giving the condition a direct and repeated influence on vector-field computation rather than relying only on prompt-level conditioning. Figure~\ref{fig:architecture} summarizes this estimation-to-expression pipeline.

We evaluate ALIGN using standard and retry-heavy successful real-robot demonstrations for policy training; failed trajectories are used only for held-out condition evaluation. Our contributions are threefold:
\vspace{-0.3em}
\begin{itemize}[leftmargin=*, itemsep=1pt, topsep=2pt, parsep=0pt]
  \item We formulate AC-BC for mixed-quality robot data as two coupled problems: condition construction and condition expression, diagnosed by AC-BC Condition Fidelity (ACF) and Conditional Vector-Field Separability (CVFS).
  \item We propose a direct categorical pairwise condition estimator that predicts the policy-facing AC-BC condition directly, reducing conversion error relative to value/progress thresholding.
  \item We introduce AdaRMS action-expert injection for flow-based VLA policies and validate ALIGN across two real-robot embodiments under controlled standard and retry-heavy training mixtures, showing improved condition fidelity, stronger condition-dependent vector-field separation, and higher closed-loop success.
\end{itemize}
\vspace{-0.3em}

\section{Related Work}

\noindent\textbf{Vision-language-action models for robot control.} Modern robot policies increasingly use vision-language-action (VLA) backbones that combine perception, language, robot state, and action prediction, spanning embodied multimodal models, scalable transformer policies, cross-embodiment data, open-source generalists, and foundation-model-coordinated visuomotor policies~\citep{driess2023palm,brohan2022rt,zitkovich2023rt,o2024open,team2024octo,kim2024openvla,shukor2025smolvla,cai2026xiaomi,bjorck2025gr00t,he2025falcon,team2025gemini,team2025gemini15}. Diffusion- and flow-based variants further model actions through generative dynamics~\citep{chi2025diffusion,liu2025rdt,black2024pi_0,zhang2025spatial,hu2026bagelvla}. These backbones improve action modeling but leave open how mixed-quality segments should be selected or conditioned during imitation.

\noindent\textbf{Advantage-conditioned imitation and policy extraction.} Offline RL and policy extraction use learned returns, values, rewards, or advantages to bias imitation, including AWR/AWAC, CRR/IQL, and return- or goal-conditioned sequence models~\citep{peng2019advantage,ashvin2020accelerating,wang2020critic,kostrikov2021offline,chen2021decision,emmons2021rvs}. Robot variants learn rewards or progress for data selection, online RL, policy steering, and evaluation~\citep{wagenmaker2025steering,liang2026robometer,ma2022vip,ma2023liv,sontakke2023roboclip,ayalew2025progressor,tan2025robo,chen2026topreward,chen2025sarm}; ReWiND improves policies with language-guided rewards without new demonstrations~\citep{zhang2025rewind}, and GR-RL filters progress-positive transitions before VLA specialization~\citep{li2025gr}. Other work exploits imperfect or mixed-quality data by self-supervised failure scoring, segment selection, value-guided retrials, or human recovery/correction trajectories~\citep{wu2025learning,chen2025towards,du2024err,hu2025rac}. Closest to our setting, \(\pi^*_{0.6}\) converts learned score differences into positive/negative VLA conditions~\citep{intelligence2025pi06vlalearnsexperience}; \(\chi_0\) predicts a stage-conditioned pairwise advantage before thresholding it into a binary optimality label~\citep{yu2026chi_}. ARM also uses pairwise supervision, but for advantage reward modeling rather than the local condition consumed by AC-BC~\citep{mao2026arm}. ALIGN is complementary: instead of treating value, progress, reward, or advantage as a continuous score for filtering, reweighting, or later thresholding, we directly predict the categorical AC-BC condition and study how it enters the action expert.

\noindent\textbf{Conditioning interfaces for generative action policies.} Condition estimation is distinct from condition expression. VLA policies can express advantage through prompts or special tokens~\citep{intelligence2025pi06vlalearnsexperience}, while generative policies provide lower-level conditioning through timestep modulation, feature-wise affine transforms, or conditional vector-field guidance~\citep{peebles2023scalable,perez2018film,ho2022classifier}. We compare action-expert condition-injection interfaces using downstream success and CVFS; among the tested mechanisms, AdaRMS gives the strongest vector-field separation and closed-loop performance.

\section{Method}
\label{sec:method}

\noindent\textbf{Condition-Mediated Decomposition.} We formulate AC-BC as a condition-mediated policy extraction process with two coupled stages:
\[
\begin{aligned}
c_t &= g_\phi(\mathcal{D}_{\mathrm{mix}}), 
&& \text{condition construction},\\
v_t &= v_\theta(x_t,t,o_t,c_t),
&& \text{condition expression}.
\end{aligned}
\]
Here, \(\mathcal{D}_{\mathrm{mix}}\) denotes mixed-quality offline trajectories, 
\(c_t\) is the policy-facing AC-BC condition, and \(v_\theta\) is the vector field predicted by the flow-matching action expert. 
This decomposition makes explicit two places where the conditioning signal can be weakened.
The first is \emph{condition construction}: whether mixed trajectory evidence, including retries, pauses, recoveries, and failures, is mapped to the correct discrete condition consumed by the policy. 
The second is \emph{condition expression}: whether changing this condition produces a meaningful change in the features and vector field that determine action prediction.
Under this view, successful AC-BC requires not only estimating behavior quality, but also preserving the resulting condition as an effective control variable inside the policy. 
ALIGN instantiates this principle by directly estimating categorical pairwise conditions in the policy-facing label space and injecting them into the DiT action expert through AdaRMS modulation fused with flow time.
Section~\ref{sec:experiments} evaluates these two stages using held-out condition fidelity, vector-field diagnostics, and downstream robot success.

\subsection{Direct Categorical Condition Estimation}
\label{sec:condition_estimator}

\noindent\textbf{Mapping estimator outputs to AC-BC conditions.} Many robot-learning systems estimate continuous task-quality signals~\citep{yu2026chi_,tan2025robo,liang2026robometer,chen2026topreward}, but AC-BC uses them only after conversion into a discrete policy-facing condition. For the \(\pi^*_{0.6}\)-style and \(\chi_0\)-style baselines, we follow the original advantage-threshold protocols~\citep{intelligence2025pi06vlalearnsexperience,yu2026chi_}: each method computes an advantage-style score on the local comparison \((o_{t-N},o_t)\), with \(N=50\), and thresholds that score using the 30\% rule. This shared score-to-condition conversion is the condition-construction bottleneck evaluated by ACF, because continuous-score errors near the threshold become policy-facing condition errors. Appendix~\ref{app:value_estimator_architecture} gives the full formulas, threshold construction, and implementation details.

\noindent\textbf{Direct categorical condition estimator.} We replace post-hoc thresholding of a predicted scalar score with categorical estimation in the condition space consumed by AC-BC. On standard data-collection trajectories, a locally monotone value or progress target makes the value-difference rule \(V(o_j)-V(o_i)>\tau_{\mathrm{V}}\) correspond, up to local progress scale, to a minimum forward temporal gap \(j-i>\tau\). We set this temporal-gap threshold to \(\tau=30\) frames and use it to supervise pairwise conditions before estimator inference. For a randomly sampled same-trajectory pair \((o_i,o_j)\),
\[
  \tilde{c}_{i,j}=g_{\tau}(j-i),
  \qquad
  \mathcal{L}_{\mathrm{cls}}
  =
  -\log p_\phi(\tilde{c}_{i,j}\mid o_i,o_j)
\]
where \(g_{\tau}\) maps the signed frame gap to a categorical condition. After training, we run the estimator on each trajectory to infer frame-level conditions for policy learning. For timestep \(t\), this applies the fixed local comparison used by AC-BC:
\[
  \hat{c}^{\phi}_t
  =
  \operatorname*{arg\,max}_{c\in\mathcal{C}}
  p_\phi(c\mid o_{t-N},o_t),
  \qquad N=50
\]
Thus, the threshold is used to construct categorical supervision rather than to discretize a noisy scalar prediction after inference. We use binary labels as the main AC-BC interface because they match the deployment-time imitation decision, while the three-class variant tests whether finer forward/neutral/backward granularity helps or adds condition-to-policy expressivity. Appendix~\ref{app:estimator_settings} details the design rationale and gap rules, and Appendix~\ref{app:condition_flow} gives the full train/evaluation/deployment condition flow. This pairwise supervision is related in form to ARM~\citep{mao2026arm}, but our estimator targets the local AC-BC condition rather than a global progress model for downstream weighting.

\subsection{Condition Expression in the Action Expert}
\label{sec:condition_injection}

By predicting the categorical AC-BC condition directly, the estimator above improves condition fidelity, i.e., the accuracy of the predicted policy-facing condition. However, high condition fidelity is not sufficient unless the policy expresses the condition in the features that determine the action vector field. We therefore view condition expression as a measurable property of the vector-field predictor: under matched open-loop probes, changing only \(c\) should induce condition-separable changes in \(v_\theta(x_t,t,o_t,c)\). Section~\ref{sec:exp_policy_injection} later quantifies this property with Conditional Vector-Field Separability (CVFS), a separability diagnostic for whether different injection mechanisms produce condition-dependent vector-field changes. Textual advantage prompts provide the most direct baseline, e.g., ``Advantage: positive'' or ``Advantage: negative''~\citep{intelligence2025pi06vlalearnsexperience}. For a flow-based action policy, however, this path is indirect: the condition is first processed by the language or vision-language backbone and only later reaches the action expert. We therefore treat condition injection as a progression: keep the condition in the prompt/prefix stream, move it past the VLM backbone into the action-side representation, repeat it throughout the action expert, and finally use it to modulate normalized action-embedding features. This progression is motivated by denoising-path conditioning in robot diffusion policies~\citep{chi2025diffusion} and adaptive feature modulation in generative transformers~\citep{peebles2023scalable,perez2018film}; ALIGN instantiates the strongest tested stage with AdaRMS.

\noindent\textbf{AdaRMS action-expert injection.} We use AdaRMS because the policy-facing effect of a condition is not merely the presence of a text token, but a change in the vector field computed by the action expert. The action expert already uses the flow-time embedding to select a denoising-stage-specific computation; we treat \(c\) as an additional index of that computation. We map the scalar condition label to the flow-time embedding dimension and add it to the processed flow-time embedding; the summed embedding then parameterizes the AdaRMS scale, shift, and gate at each action-expert layer.

This design gives the condition a direct and repeated path into vector-field prediction: at a fixed observation, noisy action, and flow time, changing \(c\) changes the modulation applied to action-embedding features. Compared with prompt conditioning, the condition no longer has to survive only through the language/VLM pathway. Compared with a one-time suffix embedding shift, AdaRMS affects both feature amplitude and residual update strength throughout the action expert. The experiments evaluate prompt-level and post-backbone alternatives using both CVFS and closed-loop success, with implementation details in Appendix~\ref{app:injection_details}.


\section{Experiments}
\label{sec:experiments}

Our evaluation is organized around two questions:
\vspace{-0.1em}
\begin{itemize}[leftmargin=*, itemsep=1pt, topsep=2pt, parsep=0pt]
  \item Does direct categorical estimation yield more faithful AC-BC conditions than \(\pi^*_{0.6}\)- and \(\chi_0\)-style thresholding?
  \item Given the same predicted condition, does action-expert injection improve vector-field expression and closed-loop success?
\end{itemize}
\vspace{-0.5em}

\begin{figure}[htbp]
  \centering
  \includegraphics[width=\linewidth]{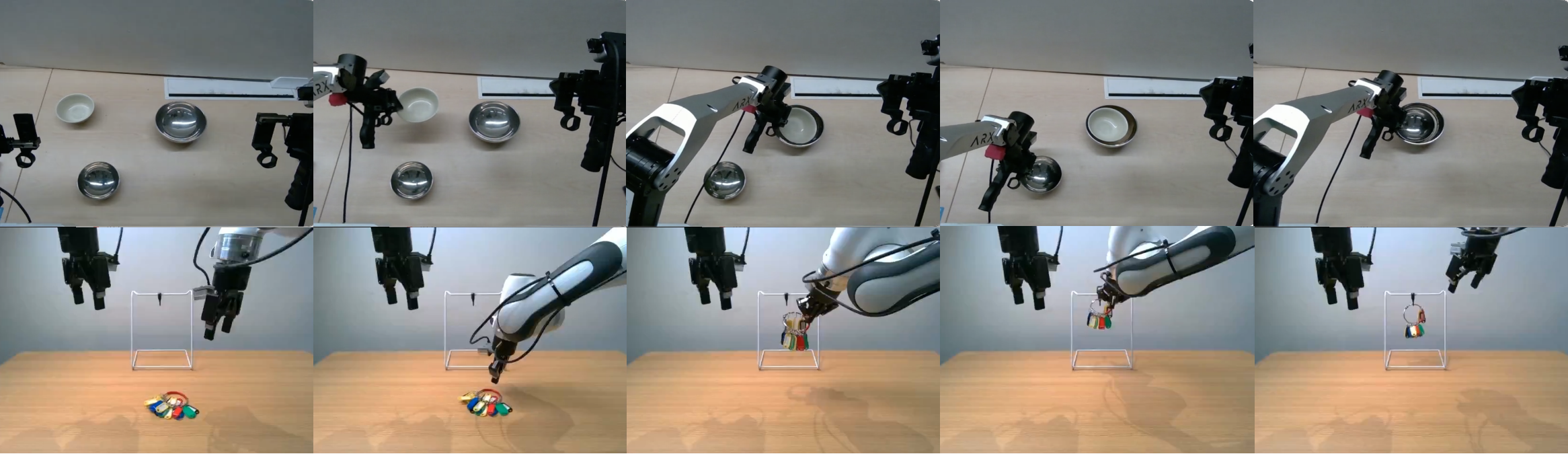}
  \vspace{-1.5em}
  \caption{
  Validation-set episodes for ARX X5 bowl stacking and Franka key hanging.
  }
  \label{fig:real_robot_tasks}
  \vspace{-1.0em}
\end{figure}

\subsection{Experimental Setting}
\label{sec:exp_setting}

We evaluate on two real-robot tasks (Figure~\ref{fig:real_robot_tasks}): ARX X5 bowl stacking and Franka key hanging. For each task, policy training mixes 100 standard trajectories with 100 retry-heavy trajectories containing both incorrect segments and recovery, testing whether AC-BC can preserve useful recovery evidence while down-conditioning low-quality frames. ACF uses task-specific held-out validation sets \(\mathcal{D}_{\mathrm{val}}\), collected and annotated for both ARX X5 and Franka, with standard, retry, and failure behavior excluded from training. Closed-loop success is measured over 30 fixed-setup rollouts per policy variant, using the same object poses, initial states, and scene setups across methods for paired comparison. Appendix~\ref{app:implementation_details} gives hardware, data, condition-flow, and hyperparameter details.

We report ACF-Precision, ACF-Recall, and ACF-F1 for estimators, closed-loop success for policies, and CVFS for condition-injection methods.

\subsection{Estimator Evaluation}
\label{sec:exp_estimator}

To answer the first question, we compare our direct estimator \(p_\phi(c\mid o_{t-N},o_t)\) with the \(\pi^*_{0.6}\)-style and \(\chi_0\)-style AC-BC baselines. For a fair ACF comparison, all estimator outputs are evaluated in the same binary AC-BC condition space: continuous-value baselines are thresholded into \(\{+,-\}\), and direct tri-class predictions are collapsed into \(\{+,-\}\).

\noindent\textbf{AC-BC Condition Fidelity.} ACF measures agreement between the induced condition \(\hat{c}_t\) and validation label \(c_t\). We report positive-class precision, recall, and F1: precision penalizes false \(+\) assignments on stagnant or regressive frames, while recall measures whether useful progress remains available for imitation. ACF is diagnostic; closed-loop success remains the policy metric. Table~\ref{tab:estimator_eval} reports these metrics for each estimator on the two held-out validation sets. The direct binary estimator achieves the highest F1 on both tasks and the highest recall, improving positive-frame recovery while maintaining high precision and the tri-class variant is close but slightly weaker.

\begin{table}[htbp]
\vspace{-0.8em}
  \centering
  \caption{Condition-estimator ACF on held-out ARX X5 and Franka validation sets.}
  \label{tab:estimator_eval}
  \resizebox{\linewidth}{!}{%
  \begin{tabular}{llcccccc}
    \toprule
    & & \multicolumn{3}{c}{ARX X5 bowl stacking} & \multicolumn{3}{c}{Franka key hanging} \\
    \cmidrule(lr){3-5}\cmidrule(lr){6-8}
    Estimator & Output space & Precision & Recall & F1 & Precision & Recall & F1 \\
    \midrule
    \(\pi^*_{0.6}\)-style value & continuous \(\rightarrow \{+,-\}\) & \textbf{98.1}\% & 71.1\% & 82.5\% & 91.4\% & 85.8\% & 88.5\% \\
    \(\chi_0\)-style stage advantage & continuous \(\rightarrow \{+,-\}\) & 97.8\% & 70.9\% & 82.3\% & \textbf{92.7}\% & 75.6\% & 83.3\% \\
    Direct tri-class & categorical \(\rightarrow \{+,-\}\) & 97.6\% & 79.0\% & 87.3\% & 90.6\% & 86.5\% & 88.5\% \\
    \textbf{Direct binary-class (ALIGN)} & \(\{+,-\}\) & 97.7\% & \textbf{79.2}\% & \textbf{87.5}\% & 90.7\% & \textbf{89.1}\% & \textbf{89.9}\% \\
    \bottomrule
  \end{tabular}
  }
  \vspace{-1.0em}
\end{table}

\begin{figure}[htbp]
  \centering
  \vspace{-0.8em}\includegraphics[width=\linewidth]{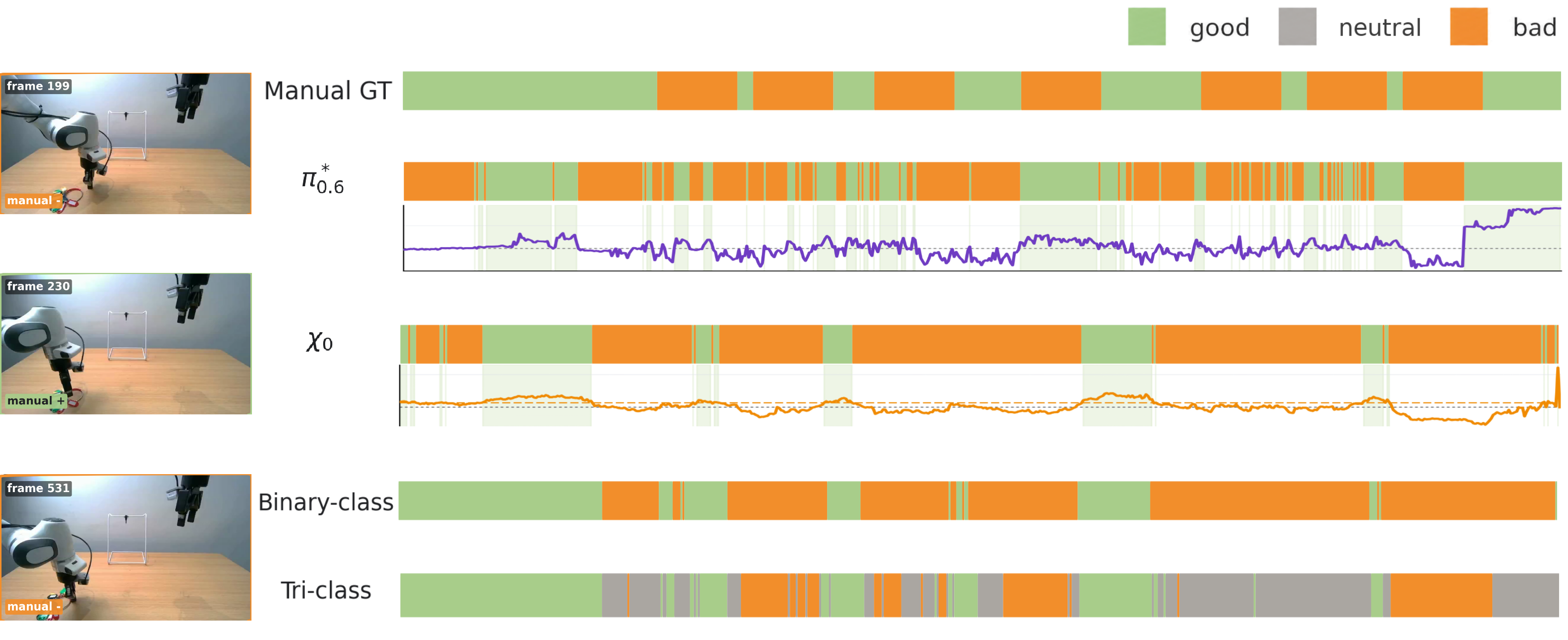}
  \vspace{-1.8em}
\caption{Estimator-induced AC-BC conditions on a held-out trajectory.}
  \vspace{-1.7em}
  \label{fig:temporal_condition_comparison}
\end{figure}

Condition construction remains imperfect for all estimators on non-monotonic robot trajectories. Figure~\ref{fig:temporal_condition_comparison} visualizes frame-level condition disagreement on a held-out trajectory: each method deviates from the validation labels(GT) at some timesteps, while the direct categorical estimator yields fewer temporal condition errors relative to thresholded continuous-score baselines. Because manual retry/failure boundaries inevitably encode annotation priors, Appendix~\ref{app:annotation_sensitivity} perturbs these boundaries and shows that the direct estimator achieves the best F1 across the tested annotation rules.

\subsection{Policy Evaluation and Injection Analysis}
\label{sec:exp_policy_injection}

To answer the second question, we isolate the effect of condition injection by fixing the condition estimator and all policy-training settings, and varying only the mechanism by which the predicted condition is fed into the policy.

\noindent\textbf{Conditional Vector-Field Separability.} Inspired by Fisher-style separability and variance-ratio criteria~\citep{fisher1936use,calinski1974dendrite}, CVFS is an open-loop diagnostic for the condition-to-policy stage. A probe is one fixed policy query, consisting of an observation \(o_i\), a noisy action \(x_{t,i}\), and a flow time \(t_i\); for each probe, we vary only the condition. An effective injection mechanism should then produce condition-specific changes in the predicted action vector field. For probe \(i\), condition \(c\in\mathcal{C}\), and \(K=|\mathcal{C}|\), define
\[
\begin{gathered}
  v_i^c=v_\theta(x_{t,i},t_i,o_i,c),
  \qquad
  \delta_i^c=v_i^c-\frac{1}{K}\sum_{c'\in\mathcal{C}}v_i^{c'},\\
  \mu_c=\frac{1}{M}\sum_{i=1}^{M}\delta_i^c,
  \qquad
  \mu=\frac{1}{K}\sum_{c\in\mathcal{C}}\mu_c
\end{gathered}
\]
where \(M\) is the number of probes. We compute between-condition and within-condition scatter,
\[
  B=\frac{1}{K-1}\sum_{c\in\mathcal{C}}\|\mu_c-\mu\|_2^2,
  \qquad
  W=\frac{1}{K(M-1)}\sum_{c\in\mathcal{C}}\sum_{i=1}^{M}\|\delta_i^c-\mu_c\|_2^2
\]
and define
\[
  \mathrm{CVFS}=\frac{B}{B+W+\epsilon}
\]
with numerical constant \(\epsilon>0\). CVFS is high when vector-field changes cluster coherently by condition, and low when the condition is ignored or expressed as inconsistent perturbations. Thus, CVFS does not replace closed-loop success; it measures whether the policy exposes a usable condition-dependent control direction before rollout.

\begin{figure}[htbp]
  \centering
  \vspace{-1.0em}
  \IfFileExists{src/figure3.pdf}{
    \includegraphics[width=\textwidth]{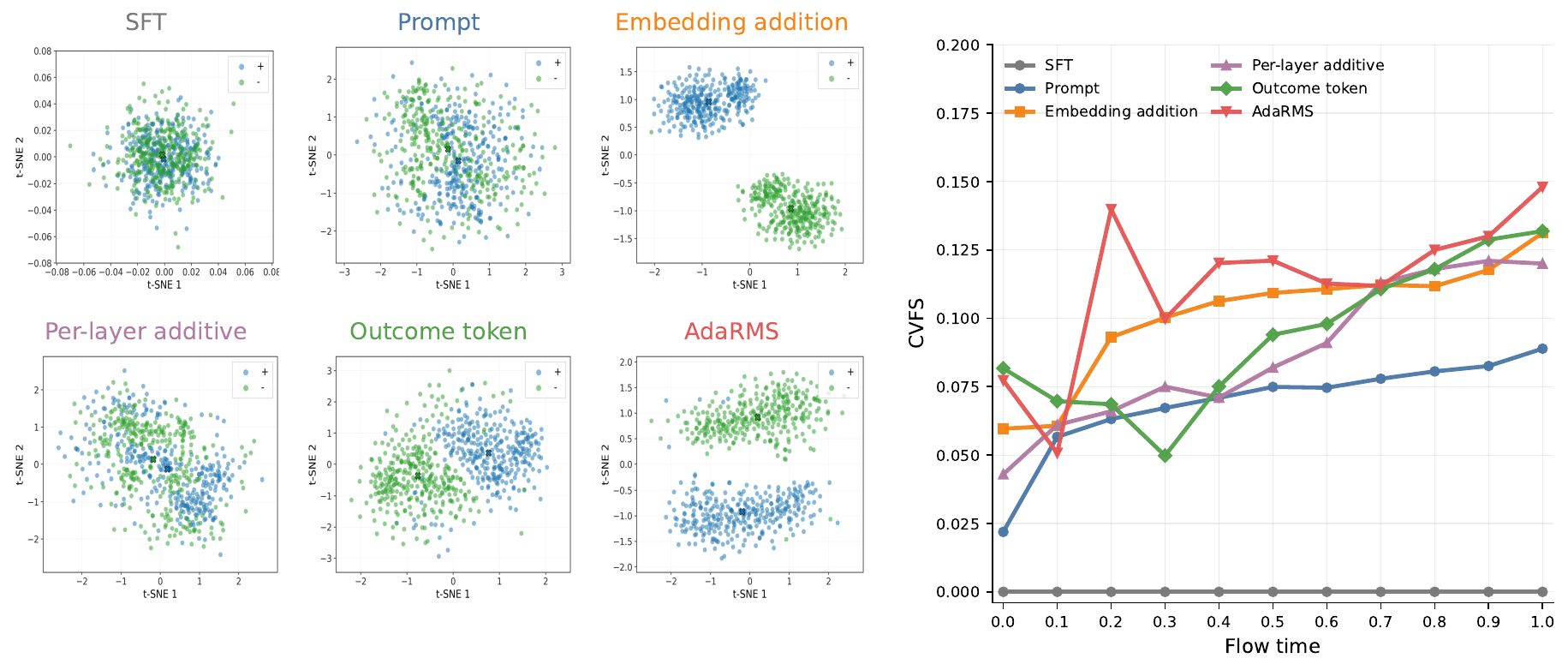}
  }{
    \fbox{
      \begin{minipage}{0.9\linewidth}
        \centering
        \vspace{0.6em}
        \textbf{Placeholder: Conditional Vector-Field Separability}
        \vspace{0.6em}

        Expected file: \texttt{src/figure3.pdf} with t-SNE panels beside the CVFS curves.
        \vspace{0.6em}
      \end{minipage}
      \vspace{-1.0em}
    }
    \vspace{-2.0em}
  }
  \vspace{-2.0em}
  \caption{Condition-dependent vector-field separation across injection mechanisms. Left: t-SNE of vector-field effects at flow time \(t=0.2\). Right: CVFS over flow time, where stronger post-backbone injection yields larger separation than prompt conditioning.}
  \label{fig:cvfs_vector_field}
  \vspace{-1.0em}
\end{figure}

\begin{figure}[!t]
  \centering
  \IfFileExists{src/figure3_injection_ablation.pdf}{
    \includegraphics[width=\textwidth]{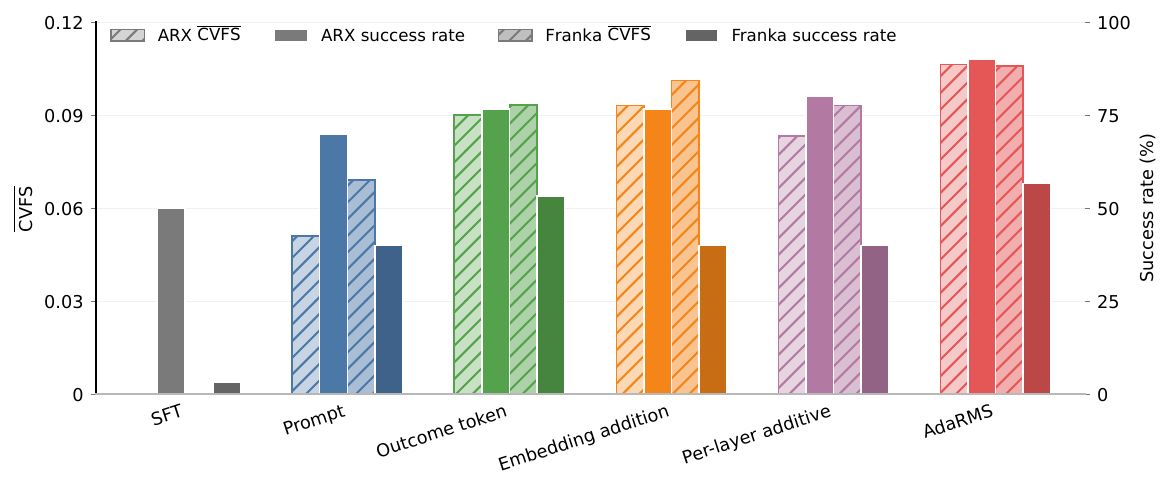}
  }{
    \fbox{
      \begin{minipage}{0.9\linewidth}
        \centering
        \vspace{1.6em}
        \textbf{Placeholder: Injection Ablation Bar Chart}
        \vspace{0.6em}

        Generate with \texttt{python scripts/make\_experiment\_assets.py} after filling \texttt{data/injection\_ablation.csv}.
        \vspace{0.6em}
      \end{minipage}
      \vspace{-2.0em}
    }
  }
  \caption{Condition-injection ablation with a fixed direct estimator. Bars report flow-time-averaged \(\overline{\mathrm{CVFS}}\) and 30-rollout closed-loop success for ARX X5 bowl stacking and Franka key hanging. Hatched bars denote \(\overline{\mathrm{CVFS}}\) on the left axis, while solid bars denote success rate on the right axis.}

  \label{fig:injection_ablation_bar}
\end{figure}

Figure~\ref{fig:cvfs_vector_field} motivates the injection study: prompt conditioning is the natural AC-BC baseline, but its vector-field separation is limited after the condition passes through the language/VLM pathway. We therefore test action-proximal injection using both CVFS and downstream success. The ablation in Figure~\ref{fig:injection_ablation_bar} suggests a positive association between flow-time-averaged CVFS and closed-loop success: methods with larger \(\overline{\mathrm{CVFS}}\) generally correspond to higher success, with AdaRMS highest on both.

\noindent\textbf{Condition-Injection Ablation.} We fix the direct estimator and vary only the injection mechanism, ordered from prompt/prefix conditioning to action-side, layer-wise, and AdaRMS modulation. This progression reflects our hypothesis that useful AC-BC conditions must be expressed inside the action expert rather than merely encoded in the prompt. It separates gains from bypassing the VLM pathway, repeated action-expert conditioning, and adaptive feature modulation; implementation details are in Appendix~\ref{app:injection_details}.

Table~\ref{tab:policy_eval} is organized around two experimental controls. The first block fixes prompt/default injection to compare condition construction. Adding retry-heavy trajectories to SFT does not produce a consistent observed improvement across the two tasks. Under the same prompt interface, the direct binary estimator reaches 70.0\% on ARX X5 and 40.0\% on Franka, with higher observed success than the thresholded baselines and the tri-class variant. The second block fixes this estimator and compares injection mechanisms. ARX X5 success rises from 70.0\% with prompts to 90.0\% with AdaRMS, and Franka success rises from 40.0\% to 56.7\%. These point estimates are consistent with contributions from both condition construction and expression, but do not establish that the two components independently improve closed-loop success.

\begin{table}[t]
  \centering
  \small
  \caption{Closed-loop policy success over 30 rollouts on ARX bowl stacking and Franka key hanging.}
  \label{tab:policy_eval}
  \resizebox{\linewidth}{!}{%
  \begin{tabular}{@{}lllcc@{}}
    \toprule
    Policy variant & Condition source & Injection & ARX bowl stacking & Franka key hanging \\
    \midrule
    \rowcolor{gray!8}
    \multicolumn{5}{@{}l}{\textit{Condition construction comparison}} \\
    SFT, 100 demonstrations & none & none & 46.7\% & 6.7\% \\
    SFT, 100 demonstrations + 100 retry trajectories & none & none & 50.0\% & 3.3\% \\
    \(\pi^*_{0.6}\)-style threshold & continuous progress & prompt/default & 50.0\% & 26.7\% \\
    \(\chi_0\)-style threshold & continuous stage advantage & prompt/default & 56.7\% & 33.3\% \\
    2-class direct estimator & binary categorical & prompt & \textbf{70.0\%} & \textbf{40.0\%} \\
    3-class direct estimator & tri-class categorical & prompt & 66.7\% & 20.0\% \\
    \midrule
    \rowcolor{gray!8}
    \multicolumn{5}{@{}l}{\textit{Condition injection comparison with the direct binary estimator}} \\
    2-class direct estimator & binary categorical & prompt & 70.0\% & 40.0\% \\
    2-class direct estimator & binary categorical & embedding addition & 76.7\% & 40.0\% \\
    2-class direct estimator & binary categorical & per-layer additive & 83.3\% & 40.0\% \\
    2-class direct estimator & binary categorical & outcome token & 76.7\% & 53.3\% \\
    \textbf{ALIGN(ours)} & binary categorical & AdaRMS & \textbf{90.0\%} & \textbf{56.7\%} \\
    \midrule
    \rowcolor{gray!8}
    \multicolumn{5}{@{}l}{\textit{Three-class condition-space extension}} \\
    3-class direct estimator & tri-class categorical & AdaRMS & 86.7\% & 33.3\% \\
    \bottomrule
  \end{tabular}%
  }
\end{table}

\noindent\textbf{Multitask Categorical Condition Injection.}\label{sec:exp_multitask_condition_injection}
The CVFS analysis above indicates that action-expert injection makes a categorical variable more explicit in the predicted vector field than prompt conditioning alone. We therefore ask whether this effect is specific to AC-BC advantage labels or reflects broader categorical steerability. We probe this question on the Franka platform in a balanced three-task setting covering bottle-cap unscrewing, bowl stacking, and key hanging: the instruction-only policy receives task-specific language prompts, while the injected policy retains the same prompts and uses AdaRMS to inject the task category through the same pathway used for advantage conditions. Matched task-category injection improves bottle-cap unscrewing from 0.0\% to 73.3\%, bowl stacking from 26.7\% to 80.0\%, and key hanging from 0.0\% to 50.0\%. Appendix~\ref{app:multitask_task_code_sweep} reports the full deployment sweep and further analysis.

\section{Limitations}

ALIGN remains task-specific: the estimator is trained within the studied manipulation tasks and does not target semantic generalization across unseen tasks, instructions, or embodiments. We train from standard data-collection trajectories with temporal-gap labels and reserve manually annotated retry/failure segments for validation to avoid leakage and task-design-specific annotation bias. Consequently, the estimator does not learn directly from labeled negative behavior; incorporating retry, failure, and non-progressive data requires label rules that control human and data-collection bias. Binary conditions cannot rank finer quality differences within the positive class. The binary/three-class comparison probes discretization of the same progress-based signal, rather than comparing binary conditioning with calibrated continuous-quality conditioning; it does not isolate label ambiguity from the policy's ability to use finer conditions. CVFS is an open-loop diagnostic: even when it correlates with success here, separable condition-dependent vector fields need not always improve behavior. Our injection study is also not exhaustive; AdaRMS is strongest among the tested post-backbone mechanisms, while other action-expert interfaces may yield more expressive conditioning.

\section{Conclusion}

We presented ALIGN, a condition-aligned AC-BC framework for extracting robust manipulation policies from mixed-quality robot data. 
ALIGN targets two key bottlenecks in VLA-based AC-BC: condition construction and condition expression. 
For condition construction, it replaces thresholded scalar value or progress differences with a direct categorical pairwise estimator that predicts the policy-facing condition. 
For condition expression, it injects the categorical condition into the DiT action expert through AdaRMS modulation fused with flow time, allowing the condition to directly influence action-vector-field prediction. 
Across two real-robot manipulation tasks with deliberately mixed standard and retry-heavy trajectories, ALIGN improves closed-loop success while also increasing condition fidelity and conditional vector-field separability. 
Within the studied tasks and temporal-gap supervision, these results support directly estimating the policy-facing categorical condition and expressing it through action-expert modulation. 
A supplementary multitask probe further suggests that CVFS and action-expert injection can expose and strengthen categorical steerability beyond advantage labels. 
Future work should broaden condition definitions and retry/failure supervision without annotation bias.

\clearpage
\acknowledgments{We thank Youchao Zhang, Youzhang He and Wenqun Wang for their insightful discussion.}

\bibliography{references}

\clearpage
\appendix
\section{Hardware Platforms, Dataset Construction, and Data Usage}
\label{app:implementation_details}
\label{app:data_details}

\subsection{Hardware Platforms}
\label{app:hardware_platforms}

\begin{figure}[H]
  \centering
  \begin{minipage}[c]{0.44\linewidth}
    \centering
    \includegraphics[width=\linewidth]{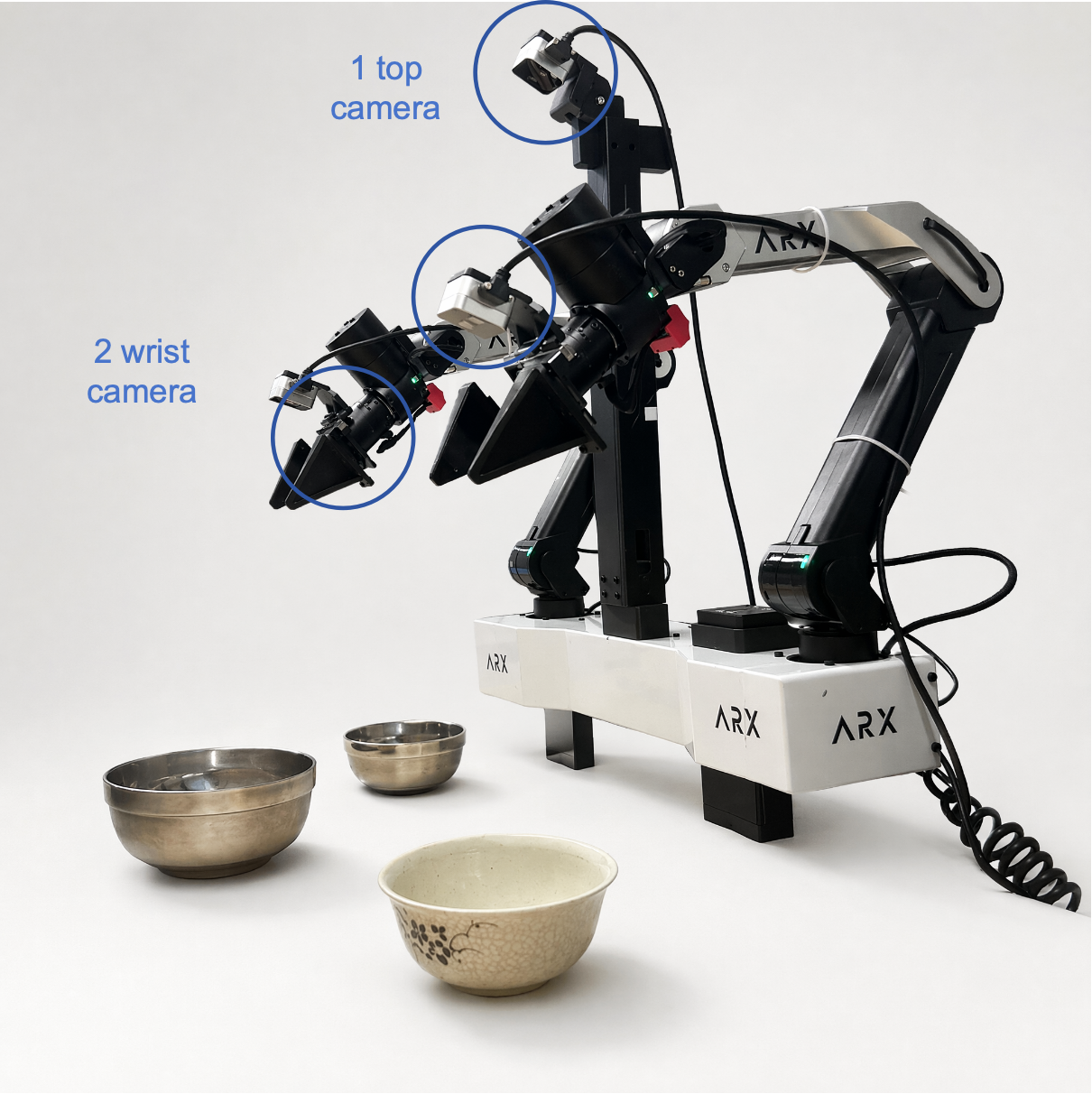}\\[-0.2em]
    \small ARX X5
  \end{minipage}
  \hfill
  \begin{minipage}[c]{0.44\linewidth}
    \centering
    \includegraphics[width=\linewidth]{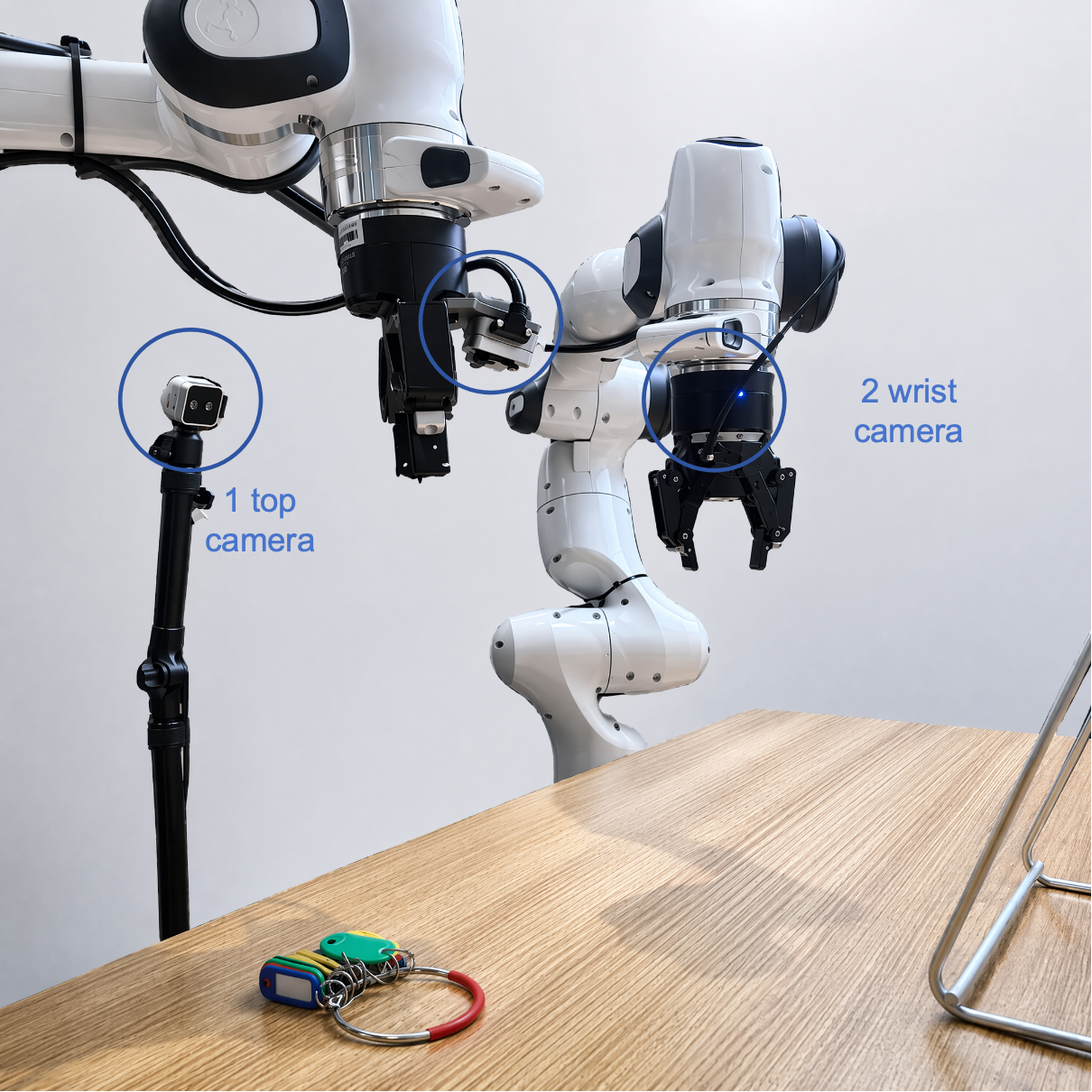}\\[-0.2em]
    \small Franka
  \end{minipage}
  \caption{Real-robot hardware platforms used in our experiments. Left: the ARX X5 platform used for bowl stacking. Right: the Franka platform used for key hanging.}
  \label{fig:hardware_photos}
\end{figure}

We evaluate ALIGN on two dual-arm tabletop manipulation platforms with different embodiments and control interfaces. Both systems provide the policy with synchronized multi-view RGB observations, robot proprioception, and a task instruction. Figure~\ref{fig:hardware_photos} shows the physical platforms, and Table~\ref{tab:hardware_platforms} summarizes their hardware interfaces.

\noindent\textbf{ARX X5.} The ARX X5 setup follows an ALOHA-style bimanual manipulation layout. It contains two 6-DoF robot arms, each equipped with a 1-DoF parallel gripper. The visual stream is collected at 30\,Hz from three RGB cameras: one front-facing camera covering the tabletop workspace and two wrist-mounted cameras providing local views near the end effectors. Robot commands use 100\,Hz position control. We use this platform for the three-bowl stacking task.

\noindent\textbf{Franka Research 3.} The Franka setup consists of two Franka Research 3 arms arranged as a dual-arm tabletop system. Each arm provides 7-DoF manipulation and is equipped with a parallel gripper. As in the ARX setting, the workspace is observed at 30\,Hz through one front-facing RGB camera and two wrist-mounted RGB cameras. Commands are issued at 200\,Hz through the Franka Control Interface under Cartesian impedance control. We use this platform for the key-hanging task.

\begin{table}[H]
  \centering
  \small
  \caption{Real-robot hardware and observation interfaces used in our experiments.}
  \label{tab:hardware_platforms}
  \resizebox{\linewidth}{!}{%
  \begin{tabular}{@{}lllll@{}}
    \toprule
    Platform & Robot embodiment & End effectors & Cameras & Control \\
    \midrule
    ARX X5 & Dual 6-DoF arms & Dual 1-DoF parallel grippers & Front RGB + two wrist RGB, 30\,Hz & 100\,Hz position control \\
    Franka Research 3 & Dual 7-DoF arms & Dual parallel grippers & Front RGB + two wrist RGB, 30\,Hz & 200\,Hz Cartesian impedance \\
    \bottomrule
  \end{tabular}%
  }
\end{table}

For policy learning, each observation \(o_t\) absorbs the multi-view images, robot state, and task instruction; the paper therefore uses the same observation notation across hardware platforms even though their proprioceptive and control interfaces differ. During closed-loop evaluation, each policy variant is tested on the same fixed task apparatus and the same suite of setup configurations within a platform, so comparisons are not confounded by method-specific changes to camera placement, robot embodiment, or scene layout.

\subsection{Robot Tasks and Data Splits}
\label{app:data_splits}

We evaluate on two real-robot manipulation tasks. The ARX bowl-stacking task uses an ARX X5 arm and requires stacking three bowls in a specified order. The Franka key-hanging task uses a Franka arm and requires placing a key onto a hook. For each task, policy training deliberately combines 100 standard data-collection trajectories with 100 retry-heavy trajectories in which retry actions are inserted by design. These retry trajectories contain useful recovery behavior together with incorrect or stagnant segments, creating mixed-quality training data for testing whether AC-BC can extract recovery evidence more selectively than SFT.

Estimator evaluation uses a separate validation set \(\mathcal{D}_{\mathrm{val}}\) that is excluded from both estimator training and policy training. For each task, \(\mathcal{D}_{\mathrm{val}}\) contains 100 additional standard data-collection trajectories, 100 additional trajectories with deliberately inserted retry actions, and 50 additional trajectories with deliberately designed failure actions. Closed-loop policy success is measured over 30 robot rollouts per policy variant. A rollout is counted as successful only if the task objective is completed without human intervention.

\subsection{Validation Annotation Protocol}
\label{app:validation_labels}

The validation set is designed to test whether an estimator induces the correct frame-level AC-BC condition under standard, retry, and failure behavior. Annotators mark the start frame \(\tau\) of each designed retry or failure event. The marked events correspond to the deliberately inserted action patterns in the data-collection protocol; therefore, the labels are tied to known intervention points rather than inferred from estimator predictions.

We use short post-event windows to define the binary validation condition. Retry events label the next \(K_r=20\) frames as negative, and failure events label the next \(K_f=30\) frames as negative:
\[
  c_t =
  \begin{cases}
  -, & t\in[\tau,\tau+K_r]\ \text{for retry events}\\
  -, & t\in[\tau,\tau+K_f]\ \text{for failure events}\\
  +, & \mathrm{otherwise}
  \end{cases}
\]
where \(+\) denotes useful imitation evidence and \(-\) denotes stagnant or regressive evidence for binary condition evaluation. When multiple annotated windows overlap, the negative label takes precedence. These labels are used only for held-out estimator evaluation and are not used to train the gap-supervised estimator.

\subsection{Annotation-Boundary Robustness of the Direct Estimator}
\label{app:annotation_sensitivity}

Frame-level retry and failure boundaries are manually specified and therefore can introduce unavoidable boundary bias. To test whether the estimator ranking is tied to one particular annotation rule, we perturb the validation labels in three ways and recompute ACF. The baseline rule is the manual window definition used in the main text. The \emph{ignore boundary} rule removes \(m\) frames on each side of each annotated boundary from evaluation, so ambiguous transition frames do not count as either positive or negative. The \emph{expand bad window} rule increases each negative retry/failure interval by \(m\) frames on both sides, making the bad-label definition more conservative. The \emph{shrink bad window} rule removes \(m\) frames from the beginning and end of each negative interval, keeping only the central part of the annotated bad segment. Tables~\ref{tab:annotation_sensitivity_arx} and~\ref{tab:annotation_sensitivity} report ACF-F1 for the ARX X5 and Franka validation sets, respectively. We report baseline precision/recall/F1 for both validation sets in Table~\ref{tab:estimator_eval}; the appendix perturbations show that the direct binary estimator remains the strongest or tied-strongest condition estimator under the tested manual-boundary variants.

\begin{table}[H]
  \centering
  \small
  \caption{Annotation-boundary robustness of ACF-F1 on the held-out ARX X5 validation set. The raw row gives the baseline manual labels reported in Table~\ref{tab:estimator_eval}; other rows perturb the manually annotated retry/failure boundaries to account for boundary bias.}
  \label{tab:annotation_sensitivity_arx}
  \resizebox{\linewidth}{!}{%
  \begin{tabular}{lccccc}
    \toprule
    Annotation rule & \(\pi^*_{0.6}\) F1 & \(\chi_0\) F1 & Binary F1 & Tri F1 & Best F1 \\
    \midrule
    Baseline manual labels & 82.5 & 82.3 & 87.5 & 87.3 & Binary \\
    Ignore boundary by 1 frame & 83.2 & 83.0 & 88.4 & 88.3 & Binary \\
    Ignore boundary by 5 frames & 83.3 & 83.2 & 88.6 & 88.4 & Binary \\
    Ignore boundary by 10 frames & 83.5 & 83.3 & 88.7 & 88.5 & Binary \\
    Expand bad window by 1 frame & 82.5 & 82.2 & 87.5 & 87.3 & Binary \\
    Expand bad window by 5 frames & 82.5 & 82.2 & 87.5 & 87.3 & Binary \\
    Expand bad window by 10 frames & 82.6 & 82.2 & 87.4 & 87.2 & Binary \\
    Shrink bad window by 1 frame & 82.4 & 82.3 & 87.5 & 87.3 & Binary \\
    Shrink bad window by 5 frames & 82.4 & 82.3 & 87.6 & 87.4 & Binary \\
    Shrink bad window by 10 frames & 82.4 & 82.3 & 87.6 & 87.5 & Binary \\
    \bottomrule
  \end{tabular}%
  }
\end{table}

\begin{table}[H]
  \centering
  \small
  \caption{Annotation-boundary robustness of ACF-F1 on the held-out Franka validation set. The main text uses the baseline manual labels; other rows perturb the manually annotated retry/failure boundaries to account for boundary bias.}
  \label{tab:annotation_sensitivity}
  \resizebox{\linewidth}{!}{%
  \begin{tabular}{lccccc}
    \toprule
    Annotation rule & \(\pi^*_{0.6}\) F1 & \(\chi_0\) F1 & Binary F1 & Tri F1 & Best F1 \\
    \midrule
    Baseline manual labels & 88.5 & 83.3 & 89.9 & 88.5 & Binary \\
    Ignore boundary by 2 frames & 89.2 & 83.9 & 89.5 & 89.2 & Binary \\
    Ignore boundary by 5 frames & 90.1 & 84.8 & 90.4 & 90.1 & Binary \\
    Ignore boundary by 10 frames & 91.7 & 86.3 & 92.0 & 91.7 & Binary \\
    Expand bad window by 2 frames & 88.4 & 83.3 & 88.7 & 88.4 & Binary \\
    Expand bad window by 5 frames & 88.3 & 83.4 & 88.5 & 88.2 & Binary \\
    Expand bad window by 10 frames & 88.1 & 83.4 & 88.1 & 87.8 & \(\pi^*_{0.6}\)/Binary \\
    Shrink bad window by 2 frames & 88.6 & 83.2 & 88.9 & 88.6 & Binary \\
    Shrink bad window by 5 frames & 88.7 & 83.1 & 89.1 & 88.8 & Binary \\
    Shrink bad window by 10 frames & 88.9 & 82.9 & 89.5 & 89.1 & Binary \\
    \bottomrule
  \end{tabular}%
  }
\end{table}

\subsection{Condition Flow}
\label{app:condition_flow}

\noindent\textbf{Algorithm 1: Condition generation and use.}

\noindent\textbf{Input:} standard trajectories \(\mathcal{D}_{\mathrm{std}}\), policy-training trajectories \(\mathcal{D}_{\mathrm{pol}}\), local horizon \(N=50\), condition vocabulary \(\mathcal{C}\), temporal-gap rule \(g_{\tau}\) with \(\tau=30\) frames.

\noindent\textbf{Output:} conditioned policy \(\pi_\theta(a\mid o,c_{\mathrm{goal}})\).
\begin{enumerate}
  \item \textbf{Estimator training.} Randomly sample two frames \((o_i,o_j)\) from the same standard data-collection trajectory in \(\mathcal{D}_{\mathrm{std}}\), without sorting them by time. Assign the weak condition label
  \[
    \tilde{c}_{i,j}=g_{\tau}(j-i)
  \]
  and train the estimator with cross-entropy,
  \[
    \mathcal{L}_{\mathrm{cls}}
    =
    -\log p_\phi(\tilde{c}_{i,j}\mid o_i,o_j)
  \]
  Deliberately collected retry and failure trajectories are not used for this gap-based estimator training.
  \item \textbf{Policy-training condition assignment.} For each frame \(t\in\mathcal{D}_{\mathrm{pol}}\), the direct estimator assigns
  \[
    \hat{c}^{\phi}_t
    =
    \operatorname*{arg\,max}_{c\in\mathcal{C}}
    p_\phi(c\mid o_{t-N},o_t)
  \]
  Value-based baselines instead compute \(\hat{c}^{V}_t\) by thresholding \(V(o_t)-V(o_{t-N})\). The induced condition is attached to the policy-training tuple and supplied through the selected prompt or injection mechanism.
  \item \textbf{Policy training.} Train the flow-matching policy by behavior cloning on conditioned tuples \((o_t,a_t,\hat{c}_t)\). All policy variants use matched data, action representation, initialization, and optimizer settings unless otherwise stated.
  \item \textbf{Closed-loop deployment.} The estimator is not run online during robot rollouts. The policy is queried with a fixed desired condition \(c_{\mathrm{goal}}\). For binary policies, \(c_{\mathrm{goal}}=+\). For multi-class policies, \(c_{\mathrm{goal}}\) is the most positive progress class.
\end{enumerate}

\subsection{How Conditions Are Used}
\label{app:condition_use}

The same condition variable has three distinct roles. During estimator training, it is a weak label derived from temporal order in standard trajectories. During policy training, it is an offline label attached to each frame and consumed by the policy through a prompt or action-expert injection mechanism. During deployment, it is not re-estimated online; instead, the policy is run under the desired successful condition. This separation avoids leakage from the validation annotations into estimator training and clarifies that the estimator is used to construct policy-training conditions, not as an online closed-loop module.

\section{Value/Progress Baselines, Model Architectures, and Training Details}
\label{app:model_training_details}

\subsection{Policy Fine-Tuning Setup}
\label{app:policy_finetuning_setup}

All policy variants are fine-tuned from the same \(\pi_{0.5}\)-style VLA checkpoint with a flow-matching action expert. The observation contains multi-view images, the robot state, and the task instruction, and the policy predicts an action chunk of horizon \(H=50\). Across variants, we keep the backbone, action representation, data split, optimizer, and fine-tuning schedule fixed; only the condition source or the condition-injection mechanism changes.

\subsection{Continuous-Score Baseline Implementation}
\label{app:value_estimator_architecture}

We treat value and progress-difference baselines under the same value-function interface. Let \(V_{\psi}(o_t)\) denote a scalar estimator trained on standard data-collection trajectories. Depending on the baseline, the regression target \(y_t\) may correspond to a value or progress annotation, and the resulting score is converted to the same AC-BC condition through a local advantage-style test. The estimator is trained with
\[
  \mathcal{L}_{V}
  =
  \left\|
  V_{\psi}(o_t)-y_t
  \right\|_2^2
\]
where \(y_t\) is the scalar supervision provided by the corresponding baseline. This keeps the comparison focused on how a continuous estimator is converted into a policy-facing condition, rather than on the particular semantics of its scalar target.

For a queried timestep \(t\), the value-based baselines compute an \(N\)-step advantage proxy using the same local comparison pair \((o_{t-N},o_t)\):
\[
  A_N(t)
  =
  R_{t-N}^{N}+V_{\psi}(o_t)-V_{\psi}(o_{t-N}),
  \qquad
  R_{t-N}^{N}=\sum_{k=t-N}^{t-1}r_k
\]
The \(\pi^*_{0.6}\)-style baseline fixes \(N=50\). We set \(\epsilon_{\mathrm{adv}}\) by applying the original 30\% advantage-threshold rule to \(A_N(t)\), without tuning it on the held-out validation labels. Since \(R_{t-N}^{N}\) is fixed, thresholding \(A_N(t)\) is equivalent to thresholding the value difference:
\[
  A_N(t)>\epsilon_{\mathrm{adv}}
  \quad\Longrightarrow\quad
  V_{\psi}(o_t)-V_{\psi}(o_{t-N})>\eta,
  \qquad
  \eta=\epsilon_{\mathrm{adv}}-R_{t-N}^{N}
\]
The induced binary AC-BC condition is therefore
\[
  \hat{c}^{V}_t =
  \begin{cases}
  +, & V_{\psi}(o_t)-V_{\psi}(o_{t-N})>\eta,\\
  -, & \mathrm{otherwise}
  \end{cases}
  \qquad N=50
\]
The \(\chi_0\)-style baseline is separated from this value-difference conversion. Following the Stage Advantage view, let \(g_i\) denote the semantic stage at the earlier frame. Instead of learning a per-frame value and subtracting two independent predictions, the stage-advantage estimator is trained on same-episode pairs to predict a pairwise advantage target \(A^\chi_{i,j}\):
\[
  \mathcal{L}_{\chi}
  =
  \left\|
  f_{\psi}(o_i,o_j,g_i)-A^\chi_{i,j}
  \right\|_2^2 .
\]
For the local AC-BC comparison used to label frame \(t\), this gives the continuous pairwise score
\[
  A^{\chi}_{\psi}(t)
  =
  f_{\psi}(o_{t-N},o_t,g_{t-N}).
\]
This avoids the value-baseline noise source \(V_{\psi}(o_t)-V_{\psi}(o_{t-N})\), but it still produces an intermediate continuous advantage score. We therefore apply the same 30\% advantage-threshold protocol to obtain the binary AC-BC interface,
\[
  \hat{c}^{\chi}_t =
  \begin{cases}
  +, & A^{\chi}_{\psi}(t)>\epsilon_{\chi},\\
  -, & \mathrm{otherwise}.
  \end{cases}
\]
The two baselines therefore threshold different advantage quantities. The value/progress baseline thresholds an \(N\)-step advantage formed from a value difference, whereas the \(\chi_0\)-style baseline thresholds the predicted pairwise stage advantage \(A^{\chi}_{\psi}(t)\) directly. ACF is computed after thresholding because the resulting \(\hat{c}_t\), not the raw scalar value, is the condition consumed by AC-BC. ALIGN removes this thresholded-score interface by training \(p_{\phi}(c\mid o_{t-N},o_t)\) directly in the categorical AC-BC condition space. Thus, Table~\ref{tab:estimator_eval} evaluates the final discrete condition consumed by AC-BC, rather than the raw continuous score before thresholding.

\subsection{Direct Categorical Estimator}
\label{app:estimator_settings}

The direct estimator receives two randomly sampled frames from the same trajectory. The sampled frames are not ordered before labeling; the label is determined by their signed temporal difference using a temporal-gap threshold of \(\tau=30\) frames. Each observation contains the same information used by the policy: images, robot state, and task instruction. In the notation, the instruction is absorbed into \(o_t\).

We consider binary and three-class variants for different reasons. The binary design is the main setting because it matches the downstream AC-BC interface: the policy ultimately needs to distinguish frames that should be imitated under the desired condition from frames that should be down-conditioned. It is therefore intentionally conservative: only pairs with sufficiently forward temporal progress are labeled positive, while reverse or near-stationary pairs are not treated as positive evidence. The three-class design is a diagnostic label-space variant. It separates clearly backward pairs from temporally ambiguous pairs, allowing us to test whether preserving a neutral middle state improves condition construction or whether the additional class mainly introduces ambiguity for policy conditioning.

For the binary estimator, the condition vocabulary is \(\mathcal{C}_2=\{+,-\}\). Let \(\Delta=j-i\). The binary temporal-gap rule is
\[
  g^{(2)}_{\tau}(\Delta)=
  \begin{cases}
  +, & \Delta>\tau,\\
  -, & \mathrm{otherwise}
  \end{cases}
\]
We use \(\tau=30\) frames in all direct-estimator experiments. Thus, only a sufficiently forward pair is labeled \(+\); reverse pairs and temporally close pairs are labeled \(-\).

For the three-class estimator, the condition vocabulary is \(\mathcal{C}_3=\{\mathrm{good},\mathrm{mid},\mathrm{bad}\}\). The label rule is
\[
  g^{(3)}_{\tau}(i,j)=
  \begin{cases}
  \mathrm{good}, & j-i>\tau,\\
  \mathrm{bad}, & i-j>\tau,\\
  \mathrm{mid}, & |j-i|\le \tau
  \end{cases}
\]
where the middle class represents temporally ambiguous pairs. The estimator is trained with cross-entropy on the corresponding label, \(\tilde{c}_{i,j}=g^{(2)}_{\tau}(j-i)\) for the binary model and \(\tilde{c}_{i,j}=g^{(3)}_{\tau}(i,j)\) for the three-class model. For binary ACF evaluation, three-class outputs are collapsed by treating \(\mathrm{good}\) as \(+\) and both \(\mathrm{mid}\) and \(\mathrm{bad}\) as \(-\).

\subsection{Condition Injection Implementations}
\label{app:injection_details}

All injection variants receive the same scalar condition label \(s_c\). For binary policies, \(s_c=1.0\) denotes \(+\) and \(s_c=0.0\) denotes \(-\). Multi-class policies use ordered scalar labels, with the most positive class queried at deployment. Let \(f_c(s_c)\) denote the condition embedding MLP, \(m_t(t)\) the processed flow-time embedding, and \(h_l\) the action-embedding hidden state at action-expert layer \(l\).

\noindent\textbf{Prompt conditioning.} Prompt conditioning appends the condition as a textual suffix to the task instruction, e.g., ``Advantage: positive'' or ``Advantage: negative''. This leaves the policy architecture unchanged, but routes the condition through the language or vision-language backbone before it can affect action-embedding features.

\noindent\textbf{Outcome-token conditioning.} Outcome-token conditioning maps \(s_c\) through the condition MLP with prefix-token hidden width and appends the resulting vector as one additional prefix token after the image and language tokens:
\[
  q_c=f^{\mathrm{tok}}_c(s_c)
\]
Suffix action embeddings can attend to this prefix token through the usual prefix-to-suffix attention pathway.

\noindent\textbf{Embedding addition.} Embedding addition maps \(s_c\) to the action-expert width and adds the resulting vector to the initial suffix action embeddings before action-expert processing. In the \(\pi_{0.5}\) implementation, each action embedding is the linear projection of the noisy action:
\[
  h_0^{a}
  =
  W_a x_t
  +
  f^{\mathrm{add}}_c(s_c)
\]
This can be viewed as prompt-style condition information placed directly on the action side after the VLM prefix.

\noindent\textbf{Per-layer additive conditioning.} Per-layer additive conditioning first combines the sinusoidal time embedding and condition embedding,
\[
  r(c,t)
  =
  f^{\mathrm{add}}_{\mathrm{time}}\!\left([\mathrm{PE}(t), f_c(s_c)]\right)
  \]
and then injects this vector at every action-expert layer through a learned layer-wise scale:
\[
  h_l'
  =
  h_l+\alpha_l\odot r(c,t)
\]
where \(\alpha_l\) is the learned scale vector for layer \(l\). This is the post-backbone additive baseline: the condition is persistent across layers, but it enters only as an additive residual.

\noindent\textbf{AdaRMS conditioning.} AdaRMS is the main injection mechanism used by ALIGN. The implementation first computes a sinusoidal flow-time embedding with periods in \([4\times10^{-3},4]\), passes it through a two-layer SiLU MLP, and adds an MLP embedding of the scalar condition label:
\[
  m_t(t)
  =
  \mathrm{SiLU}\!\left(W_t^2\,\mathrm{SiLU}(W_t^1\,\mathrm{PE}(t))\right),
  \qquad
  z(c,t)=m_t(t)+f_c(s_c)
\]
Each AdaRMS site in the action expert then maps \(z(c,t)\) to scale, shift, and gate parameters:
\[
\begin{aligned}
  \gamma_l,\beta_l,g_l&=D_l(z(c,t)),\\
  \tilde{h}_l
  &=
  (1+\gamma_l)\odot \mathrm{RMSNorm}(h_l)+\beta_l,\\
  h_{l+1}
  &=
  h_l+g_l\odot F_l(\tilde{h}_l)
\end{aligned}
\]
Here \(l\) indexes an AdaRMS site, \(D_l\) is the dense modulation layer inside the corresponding action-expert RMSNorm, and \(F_l\) denotes the associated attention or feed-forward sub-block. Thus, unlike embedding addition, AdaRMS lets the condition alter both feature normalization and residual update strength throughout vector-field prediction.

\subsection{Metric Definitions}
\label{app:metric_definitions}

ACF measures condition-level agreement on the held-out validation set after every estimator output is mapped into \(\{+,-\}\):
\[
  \mathrm{ACF\text{-}Precision}
  =
  P(c_t=+\mid \hat{c}_t=+),
  \qquad
  \mathrm{ACF\text{-}Recall}
  =
  P(\hat{c}_t=+\mid c_t=+)
\]
We also report their harmonic mean,
\[
  \mathrm{ACF\text{-}F1}
  =
  \frac{
    2\,\mathrm{ACF\text{-}Precision}\,\mathrm{ACF\text{-}Recall}
  }{
    \mathrm{ACF\text{-}Precision}+\mathrm{ACF\text{-}Recall}
  }
\]
Precision penalizes false-positive high-condition assignments on retry or failure frames, recall measures whether useful frames are preserved as positive imitation evidence, and F1 summarizes the precision--recall tradeoff.

The main text defines CVFS. In implementation, each matched probe fixes the observation, noisy action, and flow time, sweeps all conditions, and subtracts the probe-specific mean vector field before computing the scatter ratio. Unless otherwise stated, CVFS is evaluated over \(t\in\{0,0.1,0.2,\ldots,0.9,1.0\}\) and averaged over flow time for the bar chart.

\subsection{Training Hyperparameters}
\label{app:training_hyperparameters}

\begin{table}[H]
  \centering
  \caption{Policy fine-tuning settings.}
\label{tab:policy_hyperparams}
  \begin{tabular}{p{0.34\linewidth}p{0.58\linewidth}}
    \toprule
    Hyperparameter & Value \\
    \midrule
    Policy backbone & \(\pi_{0.5}\)-style VLA policy with a flow-matching action expert \\
    Action horizon  & 50  \\
    Training precision & bfloat16 \\
    Optimizer & AdamW \\
    AdamW betas / epsilon & \((0.9,0.95)\) / \(10^{-8}\) \\
    Weight decay & \(10^{-10}\) \\
    Learning-rate schedule & cosine decay \\
    Peak / final learning rate & \(2.5\times 10^{-5}\) / \(2.5\times 10^{-6}\) \\
    Batch size & 256 \\
    Training steps & 5k for conditioned policy fine-tuning \\
    Default inference denoising steps & 10 Euler steps \\
    \bottomrule
  \end{tabular}
  \vspace{-1.0em}
\end{table}

\begin{table}[H]
  \centering
  \caption{ALIGN and \(\pi^*_{0.6}\)-style estimator settings.}
  \label{tab:estimator_hyperparams}
  \begin{tabular}{p{0.34\linewidth}p{0.58\linewidth}}
    \toprule
    Setting & Value \\
    \midrule
    Pair sampling & randomly sample two frames from the same trajectory without sorting their temporal order \\
    Direct-estimator objective & cross-entropy on binary \(g^{(2)}_{\tau}(j-i)\) or three-class \(g^{(3)}_{\tau}(i,j)\) labels \\
    Value/progress objective & MSE regression for \(V_\psi(o_t)\) on scalar value or progress target \\

    Local comparison horizon & \(N=50\) \\
    Retry validation window & \(K_r=20\) frames \\
    Failure validation window & \(K_f=30\) frames \\
    Binary injection label values & \(+=1.0,\ -=0.0\) \\
    Positive-label rule & rank advantage scores; label the top 70\% as positive and the rest as negative \\
    Three-class injection label values & \(\mathrm{good}=1.0,\ \mathrm{mid}=0.8,\ \mathrm{bad}=0.0\) \\
    ALIGN/\(\pi^*_{0.6}\)-style estimator optimizer & AdamW, \((\beta_1,\beta_2)=(0.9,0.95)\), \(\epsilon=10^{-8}\) \\
    ALIGN/\(\pi^*_{0.6}\)-style estimator learning rate & cosine decay, \(2.5\times10^{-5}\rightarrow2.5\times10^{-6}\) \\
    ALIGN/\(\pi^*_{0.6}\)-style estimator training steps & 50k \\
    \bottomrule
  \end{tabular}
\end{table}

\begin{table}[H]
  \centering
  \caption{Matched \(\chi_0\)-style estimator settings. }
  \label{tab:chi0_reproduction_settings}
  \begin{tabular}{p{0.34\linewidth}p{0.58\linewidth}}
    \toprule
    Setting & Value used in our reproduction \\
    \midrule

    Pair sampling for SA & randomly sample two frames from the same trajectory without sorting their temporal order \\
    SA supervision target & pairwise stage-advantage target \(A^\chi_{i,j}\) \\
    Stage annotation & semantic sub-goals; 4 stages for ARX X5 bowl stacking and 2 stages for Franka key hanging \\
    Positive-label rule & rank local stage-advantage scores; label the top 30\% as positive and the rest as negative \\
    Local comparison horizon & \(N=50\), same as ALIGN and \(\pi^*_{0.6}\)-style baselines \\
    Optimizer & AdamW, \((\beta_1,\beta_2)=(0.9,0.95)\), \(\epsilon=10^{-8}\) \\
    Learning rate & cosine decay, \(2.5\times10^{-5}\rightarrow2.5\times10^{-6}\) \\
    Training steps & 50k \\
    \bottomrule
  \end{tabular}
\end{table}

\section{Additional Multitask Condition-Injection Analysis}
\label{app:multitask_analysis}

\subsection{Multitask Setup}
\label{app:multitask_setup}

The main experiments use CVFS to show that action-expert injection increases the vector-field separability induced by a categorical AC-BC condition. As a supplementary behavioral test of this steerability signal, we consider a multitask setting in which the injected categorical variable represents task identity rather than an estimated advantage condition. Task identity is one concrete categorical control variable; this setting asks whether the same injection pathway can make it behaviorally salient beyond the task-specific language prompt.

The multitask policy is trained and evaluated on the Franka platform, using 200 trajectories per task for bottle-cap unscrewing, bowl stacking, and key hanging. Both the prompt-only baseline and the injected policy retain task-specific language instructions. The injected policy additionally conditions on the task category through AdaRMS, using the same categorical-condition embedding and action-expert modulation path as the advantage-conditioned policy. During evaluation, the injected policy uses the matched category for the queried task.

\subsection{Multitask Categorical Condition Injection}
\label{app:multitask_task_code_sweep}

\begin{table}[H]
  \centering
  \small
  \caption{Multitask categorical-condition injection on the Franka platform. Each row reports closed-loop success on the evaluated physical task. Prompt-only uses only the task-specific language prompt, while AdaRMS uses the same prompt together with the matched task category.}
  \label{tab:multitask_code_sweep}
  \begin{tabular}{@{}lcc@{}}
    \toprule
    Task & Prompt-only success & AdaRMS success \\
    \midrule
    Unscrew bottle cap & 0.0\% & 73.3\% \\
    Stack bowls & 26.7\% & 80.0\% \\
    Hang key & 0.0\% & 50.0\% \\
    \bottomrule
  \end{tabular}
\end{table}

Table~\ref{tab:multitask_code_sweep} reports the matched-category evaluation for each physical task. On matched open-loop multitask probes, AdaRMS increases \(\overline{\mathrm{CVFS}}\) from 0.1247 under prompt-only conditioning to 0.4032, indicating that the injected category is more clearly represented in the action vector field. The accompanying success-rate gains suggest that CVFS is aligned with whether the policy learns to use the categorical signal for task-specific behavior.

\clearpage
\section{Additional Experiments and Data Details}
\label{app:additional_controls_and_scope}

\subsection{Additional Controls on AGI G2}
\label{app:g2_controls}

We evaluate an additional open-wardrobe-door task on AGI G2, with two sub-goals: engaging the handle and opening the door. Table~\ref{tab:g2_additional_controls} reports results over 50 rollouts per method.

The Filtered-SFT control uses the same classifier to discard predicted-negative frames, which comprise 5.7\% in this setting. Filtering can retain recovery frames predicted as positive, whereas AC-BC conditions on both positive and negative examples. The \(\pi^*_{0.6}\)-style estimator with AdaRMS provides a separate control that changes the estimator while holding AdaRMS injection fixed.

\begin{table}[htbp]
  \centering
  \caption{Additional AGI G2 results, with 50 rollouts per method. The first sub-goal is engaging the handle. The final column is the mean time-to-success over successful rollouts only.}
  \label{tab:g2_additional_controls}
  \begin{tabular}{@{}lrrr@{}}
    \toprule
    Method & \shortstack{Success\\(\%)} & \shortstack{First sub-goal\\(\%)} & \shortstack{Mean time\\(s)} \\
    \midrule
    Demo & 28.0 & 36.0 & 50.2 \\
    Demo + retry & 34.0 & 46.0 & 51.7 \\
    \(\pi^*_{0.6}\) thresh. + AdaRMS & 64.0 & 70.0 & 42.5 \\
    Filtered-SFT & 76.0 & 86.0 & 50.5 \\
    ALIGN & 86.0 & 90.0 & 43.2 \\
    \bottomrule
  \end{tabular}
\end{table}

\subsection{Single-Task Data Composition}
\label{app:single_task_data_clarification}

All policy-training trajectories terminate successfully. The retry-heavy trajectories are collected end-to-end rather than spliced, with one or two deliberate retries followed by recovery. Table~\ref{tab:additional_data_splits} summarizes the data pools for each single-task setting. All condition estimators use the same additional standard trajectories, disjoint from policy-training and held-out ACF data. The held-out ACF set is also excluded from policy training; failed trajectories appear only in this evaluation set. Across the three single-task datasets, the fraction of predicted-negative frames ranges from 4.0\% to 5.7\%, with 5.7\% on G2. The collected data include natural placement, lighting, and collector variation.

\begin{table}[htbp]
  \centering
  \small
  \caption{Trajectory counts per single-task setting. The estimator-training, policy-training, and held-out ACF pools are disjoint.}
  \label{tab:additional_data_splits}
  \begin{tabular}{@{}llr@{}}
    \toprule
    Data pool & Composition & Total \\
    \midrule
    Estimator training & 400 standard & 400 \\
    Policy training & 100 standard + 100 retry-heavy & 200 \\
    Held-out ACF & 100 standard + 100 retry + 50 failure & 250 \\
    \bottomrule
  \end{tabular}
\end{table}

\subsection{Uncertainty of the Main Policy Results}
\label{app:policy_uncertainty}

Table~\ref{tab:policy_eval_wilson} adds success counts and 95\% Wilson intervals to every entry in Table~\ref{tab:policy_eval}. The construction and expression comparisons retain the point-estimate interpretation stated in the main text.

\begin{table}[htbp]
  \centering
  \small
  \caption{Success counts and 95\% Wilson score intervals for every policy entry in Table~\ref{tab:policy_eval}, with \(n=30\) rollouts per policy and task. Each cell reports \(k/n\); success percentage [95\% interval], with interval endpoints also in percent. SFT, standard uses 100 standard trajectories; SFT, standard + retry uses 100 standard and 100 retry-heavy trajectories. These are marginal intervals for each policy, not confidence intervals for paired method differences.}
  \label{tab:policy_eval_wilson}
  \setlength{\tabcolsep}{3pt}
  \begin{tabular}{@{}p{0.32\linewidth}p{0.32\linewidth}p{0.32\linewidth}@{}}
    \toprule
    Policy & ARX X5 & Franka \\
    \midrule
    \multicolumn{3}{@{}l}{\textit{Condition construction comparison}} \\
    SFT, standard & 14/30; 46.7 [30.2, 63.9] & 2/30; 6.7 [1.8, 21.3] \\
    SFT, standard + retry & 15/30; 50.0 [33.2, 66.8] & 1/30; 3.3 [0.6, 16.7] \\
    \(\pi^*_{0.6}\), prompt/default & 15/30; 50.0 [33.2, 66.8] & 8/30; 26.7 [14.2, 44.4] \\
    \(\chi_0\), prompt/default & 17/30; 56.7 [39.2, 72.6] & 10/30; 33.3 [19.2, 51.2] \\
    Direct binary, prompt & 21/30; 70.0 [52.1, 83.3] & 12/30; 40.0 [24.6, 57.7] \\
    Direct tri-class, prompt & 20/30; 66.7 [48.8, 80.8] & 6/30; 20.0 [9.5, 37.3] \\
    \midrule
    \multicolumn{3}{@{}l}{\textit{Condition injection comparison with the direct binary estimator}} \\
    Direct binary, prompt & 21/30; 70.0 [52.1, 83.3] & 12/30; 40.0 [24.6, 57.7] \\
    Direct binary, embedding addition & 23/30; 76.7 [59.1, 88.2] & 12/30; 40.0 [24.6, 57.7] \\
    Direct binary, per-layer additive & 25/30; 83.3 [66.4, 92.7] & 12/30; 40.0 [24.6, 57.7] \\
    Direct binary, outcome token & 23/30; 76.7 [59.1, 88.2] & 16/30; 53.3 [36.1, 69.8] \\
    ALIGN, AdaRMS & 27/30; 90.0 [74.4, 96.5] & 17/30; 56.7 [39.2, 72.6] \\
    \midrule
    \multicolumn{3}{@{}l}{\textit{Three-class condition-space extension}} \\
    Direct tri-class, AdaRMS & 26/30; 86.7 [70.3, 94.7] & 10/30; 33.3 [19.2, 51.2] \\
    \bottomrule
  \end{tabular}
\end{table}

\subsection{Condition Granularity}
\label{app:condition_granularity_results}

Table~\ref{tab:condition_granularity_results} compares binary, three-class, and four-class conditions. Finer discretization of the same signal does not improve these point estimates. This comparison does not test continuous conditioning under calibrated graded-quality supervision. Within the positive class, binary labels do not rank marginal recovery against clean expert motion; recovering a calibrated continuous value lies outside the estimator objective and would require additional supervision.

\begin{table}[htbp]
  \centering
  \caption{Closed-loop success for different condition granularities. Four-class results are reported for prompt injection only.}
  \label{tab:condition_granularity_results}
  \begin{tabular}{@{}llrr@{}}
    \toprule
    Injection & Condition space & ARX X5 (\%) & Franka (\%) \\
    \midrule
    Prompt & Binary & 70.0 & 40.0 \\
    Prompt & Three-class & 66.7 & 20.0 \\
    Prompt & Four-class & 66.7 & 16.7 \\
    \midrule
    AdaRMS & Binary & 90.0 & 56.7 \\
    AdaRMS & Three-class & 86.7 & 33.3 \\
    \bottomrule
  \end{tabular}
\end{table}

\subsection{Additional Diagnostics}
\label{app:additional_diagnostics}

In an exploratory analysis across the six condition-injection variants in Figure~\ref{fig:injection_ablation_bar}, CVFS and success are positively correlated: ARX has \(r=0.95\), \(p=0.003\), and Franka has \(r=0.93\), \(p=0.007\), using two-sided Pearson tests. ACF and CVFS remain mechanism diagnostics rather than substitutes for closed-loop uncertainty.

Ten additional G2 distractor trajectories, comprising five standard and five retry trajectories, yield an ACF-F1 of 87.1\%, compared with 87.6\% on the standard G2 held-out set. This check evaluates the estimator's condition fidelity on these additional trajectories.

\subsection{Scope of the Evidence}
\label{app:additional_scope}

Supervision remains a distinct design axis. Our temporal-gap labels encode local task progress, and ACF measures task-specific condition fidelity; neither establishes general-purpose semantic action-quality or calibrated value estimation.

Flow matching can represent a multimodal \(p(a\mid o,c)\) in principle, so sharing a condition need not force deterministic averaging. However, mode preservation at matched observations remains unverified. Joint multitask and mixed-quality training, interacting steering signals, and policy classes beyond the single flow-matching VLA backbone studied here remain untested.

\end{document}